\documentclass[conference]{IEEEtran}
\IEEEoverridecommandlockouts
\usepackage{cite}
\usepackage{url}

\usepackage{amsmath,amssymb,amsfonts}
\usepackage{graphicx}
\usepackage{textcomp}
\usepackage[hidelinks]{hyperref}
\usepackage{xcolor}
\usepackage{booktabs}
\usepackage{longtable}
\usepackage{multirow}

\definecolor{bestbg}{RGB}{218,242,226}
\definecolor{goodbg}{RGB}{230,241,250}
\definecolor{midbg}{RGB}{250,244,218}
\definecolor{lowbg}{RGB}{245,245,245}

\newcommand{\best}[1]{\colorbox{bestbg}{\textbf{#1}}}
\newcommand{\good}[1]{\colorbox{goodbg}{#1}}
\newcommand{\midval}[1]{\colorbox{midbg}{#1}}
\newcommand{\low}[1]{\colorbox{lowbg}{#1}}

\usepackage[table]{xcolor}
\usepackage{tabularx}
\usepackage{array}

\definecolor{nllbg}{RGB}{220,242,226}
\definecolor{rawbg}{RGB}{230,241,250}
\definecolor{gptbg}{RGB}{250,244,218}

\newcommand{\nllbest}[1]{\cellcolor{nllbg}\textbf{#1}}
\newcommand{\rawbest}[1]{\cellcolor{rawbg}\textbf{#1}}
\newcommand{\gptbest}[1]{\cellcolor{gptbg}\textbf{#1}}

\newcolumntype{Y}{>{\centering\arraybackslash}X}

\usepackage{balance}
\usepackage[most]{tcolorbox}

\newtcolorbox{findingbox}{
    colback=green!5,
    colframe=green!70!black,
    boxrule=0.8pt,
    arc=2mm,
    left=2mm,
    right=2mm,
    top=1mm,
    bottom=1mm,
    title=\textbf{Key Finding},
    fonttitle=\bfseries,
}
\usepackage{algorithm}
\usepackage{algpseudocode}
\usepackage{float}

\newtcolorbox{findingbox2}{
    colback=blue!5,
    colframe=blue!60!black,
    boxrule=0.8pt,
    arc=2mm,
    left=2mm,
    right=2mm,
    top=1mm,
    bottom=1mm,
    title=\textbf{Prompt Template},
    fonttitle=\bfseries,
}
\definecolor{lightgreen}{RGB}{240, 248, 240}
\definecolor{boxbg}{RGB}{247, 252, 247}
\definecolor{boxframe}{RGB}{120, 165, 120}
\definecolor{titlebg}{RGB}{225, 240, 225}

\usepackage{framed}
\usepackage{xcolor}

\renewenvironment{shaded}{%
  \MakeFramed{\advance\hsize-\width \FrameRestore}}%
 {\endMakeFramed}

\usepackage{array} 
\def\BibTeX{{\rm B\kern-.05em{\sc i\kern-.025em b}\kern-.08em
    T\kern-.1667em\lower.7ex\hbox{E}\kern-.125emX}}

\begin{document}
\title{\huge {\textit{Do LLMs Know a Good Hypothesis When They See One?} Logit-Based Energy Scoring Outperforms Prompted LLM-as-Judge for Scientific Hypothesis Ranking}}


\author{
\IEEEauthorblockN{Swati Rajwal\textsuperscript{\dag}\IEEEauthorrefmark{1}, Sanjay Das\textsuperscript{\dag}\IEEEauthorrefmark{1}, Tirthankar Ghosal\textsuperscript{\dag}}
\IEEEauthorblockA{\textsuperscript{\dag}
\texttt{Oak Ridge National Laboratory, \{rajwals,dass3,ghosalt\}@ornl.gov}}
 \IEEEauthorrefmark{1}These authors contributed equally to this work.
}
\renewcommand{\footnoterule}{%
  \kern -3pt
\hrule width \columnwidth
\kern 2.6pt
}
\maketitle
\begingroup
\renewcommand\thefootnote{}
\footnotetext{{\scriptsize 
This manuscript has been authored by UT-Battelle, LLC, under contract DE-AC05-00OR22725 with the US Department of Energy (DOE). The US government retains and the publisher, by accepting the article for publication, acknowledges that the US government retains a nonexclusive, paid-up, irrevocable, worldwide license to publish or reproduce the published form of this manuscript, or allow others to do so, for US government purposes. DOE will provide public access to these results of federally sponsored research in accordance with the DOE Public Access Plan (\url{https://www.energy.gov/doe-public-access-plan}).
}
}

\addtocounter{footnote}{-1}
\endgroup
\begin{abstract}
Large language models (LLMs) are increasingly used for scientific hypothesis generation. However, evaluating generated hypotheses remains a challenge for trustworthy AI-enabled scientific workflows. Existing approaches often use LLMs as judges or rely on semantic similarity, which can favor familiar ideas over novel ones. We propose a logit-based energy scoring method that evaluates hypotheses using a language model’s intrinsic confidence rather than comparative judgment. We benchmarked seven language models on 1,323 papers across 12 disciplines. Each paper was paired with its hypothesis and fifteen incorrect alternatives. Intrinsic scoring reached 33.0\% Hit@1 pooled across both scorers, compared with 16.6\% for prompted listwise ranking. The strongest configuration, a 1-billion-parameter model using logit-based energy scoring, reached 53.1\%, though this was the maximum across 14 model-by-scorer combinations selected post hoc. Overall, intrinsic model confidence shows potential for scientific hypothesis evaluation. This study also motivates future research on confidence-based methods for trustworthy AI-enabled scientific discovery.
\end{abstract}


\section{Introduction}
Recent advances in large language models (LLMs) have transformed the landscape of scientific research assistance, enabling unprecedented capabilities in hypothesis generation, literature analysis, and even autonomous discovery workflows. Pioneering systems, such as Google's AI Co-Scientist or Sakana AI, exemplify this technological trend, illustrating how LLMs can propose plausible hypotheses either in collaboration with human experts or as independent agents within scientific pipelines. These developments open new possibilities for accelerating discovery, fostering interdisciplinarity, and automating hypothesis-driven exploration in both established and emerging fields.

\begin{figure*}
    \centering
    \includegraphics[width=0.9\linewidth]{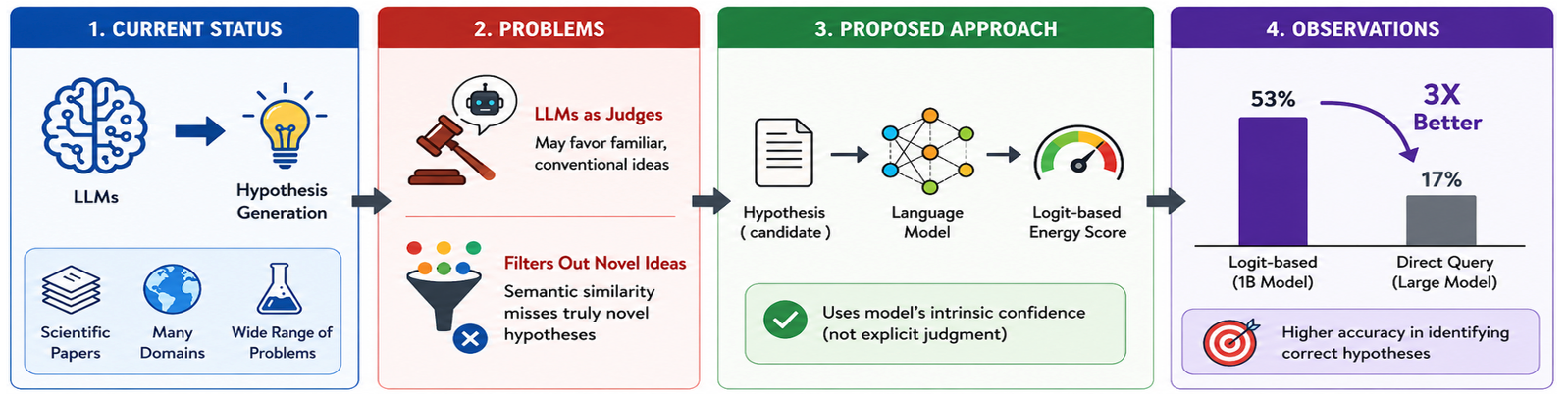}
    \caption{Overview of the study. Given a background and research question, a candidate hypothesis is scored by a language model's internal confidence rather than by prompted comparison. 53\% is the best of 14 model-by-scorer combinations.}
    \label{fig:overview}
\end{figure*}
 
A central step in the scientific process is the formulation of a hypothesis that is both consistent with prior literature and capable of explaining or predicting a phenomenon of interest. As LLMs are increasingly proposed as tools for accelerating scientific discovery (from literature-grounded ideation to automated experimental design) it becomes important to ask a more basic question: given the background and research question that motivate a study, can a language model recognize which of several candidate hypotheses is the one that domain experts actually pursued? This question is distinct from, and arguably prerequisite to, the harder task of generating novel hypotheses: before a model can be trusted to propose new scientific ideas, it should first be able to discriminate correct hypotheses from superficially plausible but incorrect ones.

The evaluation of LLM-generated scientific hypotheses currently relies primarily on \textbf{LLM-as-a-judge} frameworks, which depend on explicit verbal feedback and semantic similarity metrics, or manual human expert verification. Both paradigms pose significant constraints: automated judges frequently suffer from position bias, verbosity bias, and semantic alignment tendencies that favor conventional ideas \cite{shi-etal-2025-judging, saito2023verbositybiaspreferencelabeling, wataoka2025selfpreferencebiasllmasajudge} while discarding genuinely novel hypotheses, whereas expert human review is inherently unscalable, time-consuming, and resource-intensive.
\begin{findingbox}
A model's own likelihood over a candidate hypothesis is a stronger signal of scientific correctness than explicitly prompting a proprietary model to rank candidates. Nearly every open-weight LLM we tested beats the prompted baseline. Which intrinsic scorer is better, however, is model-dependent: the raw-logit energy criterion helps some models substantially and fails for others, and understanding why is an open question.\end{findingbox}
To overcome these limitations, we propose \textbf{Logit-based Energy Scoring}, an alternative model-intrinsic evaluation framework. By leveraging output logit probabilities as a direct proxy for internal model confidence, our approach calculates ``energy scores'' to reliably identify the most plausible candidate hypothesis without relying on verbalized judgment. Extensive benchmarking across a curated dataset reveals that this mechanism scales effectively from 1B-parameter models up to 20B systems and top proprietary baselines. Notably, lightweight open-source models such as Llama 3.2 1B match or surpass a high-capacity proprietary baseline under either intrinsic criterion, with the raw target-logit energy variant producing the largest single gain. We emphasize that the energy criterion is not uniformly superior to likelihood scoring: its benefit is concentrated in a subset of models, and characterizing that subset is a contribution of this work rather than a caveat to it. Ultimately, this work shows the potential of intrinsic confidence metrics for scientific discovery and offers a path toward democratizing advanced hypothesis evaluation via open-source models.
 
Our contributions are as follows:
\begin{enumerate}
\item This work introduces a novel, fully automatic evaluation metric for LLM-generated scientific hypotheses based on logit-derived energy scoring, thus harnessing the internal confidence distributions of language models, in contrast to conventional judge-style approaches relying on externalized judgments or semantic similarity measures.
\item The paper presents the first systematic comparison of a range of open-source LLMs and a state-of-the-art closed commercial LLM, evaluating their capabilities to discern correct scientific hypotheses, thereby highlighting performance disparities between model architectures, model scale, and evaluation paradigms.
\item Our analysis demonstrates that intrinsic model confidence provides a more reliable signal of scientific hypothesis validity than explicit LLM-judge prompting.
\end{enumerate}
The rest of the article is organized as follows. Section II
provides the background on LLM-based scientific hypothesis generation and evaluations.
Section III describes the proposed techniques. Section IV delineates the experimental setup and Section V
showcases the results. Lastly, the limitations and future works are discussed in Section VI and the article is concluded in Section VII.
\section{Background and Related Work}
\subsection{LLM-based Scientific Hypothesis Generation}
A growing body of work deploys LLMs as active participants in the scientific discovery process rather than as passive writing assistants. Google's AI co-scientist system orchestrates multiple LLM agents in a generate-debate-rank loop, producing biomedical hypotheses that were subsequently evaluated favorably by domain experts on a small number of case studies\cite{gottweis2026accelerating}. Sakana AI's AI Scientist pursues fully automated, closed-loop research: an LLM ideates a project, writes and executes code to test it, and drafts the resulting paper\cite{lu2026towards}. Related systems target materials discovery, retrosynthesis planning, and automated theorem or conjecture generation. Across this literature, hypothesis generation itself is treated as largely solved in the sense that LLMs reliably produce fluent, domain-appropriate candidate hypotheses; the harder and less standardized problem, which motivates this paper, is deciding which of many plausible-sounding candidates are actually correct or promising.

\subsection{Evaluation via LLM-as-Judge and Semantic Similarity}
The two evaluation strategies in current practice are (i) LLM-as-judge protocols, in which a (typically large, closed) model is prompted to compare a candidate hypothesis to a reference or to other candidates and produce a score, ranking, or preference\cite{NEURIPS2023_91f18a12}; and (ii) embedding-based semantic similarity, in which a candidate hypothesis and a reference hypothesis are each encoded and compared via cosine similarity or a related metric\cite{reimers-gurevych-2019-sentence}. Both strategies have been widely adopted because they are inexpensive, automatable, and correlate reasonably well with human judgment in aggregate benchmarks composed largely of previously documented findings. However, both strategies presuppose that correctness and similarity-to-reference are tightly coupled. This coupling is a much weaker assumption for open-ended or forward-looking scientific hypotheses, where a correct hypothesis is, by definition, not already documented in a form the evaluator has seen, and where surface-level plausibility can be a poor proxy for mechanistic correctness.

\subsection{Confidence and Likelihood as Evaluation Signals}
A separate line of work outside the scientific-discovery literature has studied a language model's own token-level probabilities as a signal for downstream selection tasks, including re-ranking of generated text, factuality estimation, and out-of-distribution or hallucination detection\cite{kadavath2022languagemodelsmostlyknow}. Energy-based models (EBMs) formalize an unnormalized notion of compatibility between an input and a candidate output via a scalar energy function, with lower energy corresponding to higher compatibility, and have been used to reinterpret the output layer of a classifier or language model without requiring the softmax normalization used to obtain a proper probability distribution\cite{lecun2006energy, Grathwohl2020Your, 3495724.3497526}. This literature motivates our use of raw, pre-softmax logit magnitude as a complementary signal to normalized log-likelihood: because normalization is computed over the full vocabulary at each position, it can compress or distort the very confidence signal that distinguishes a well-supported hypothesis continuation from a merely fluent one, particularly in smaller or less calibrated models.

\section{Methodology}
In this section, we present our approach to investigating whether logit-based models can reliably evaluate scientific hypotheses without relying on comparative judging. Rather than exposing the model to surface fluency or stylistic biases through comparative selection, we measure its per-token confidence given the background context. A model assigns high probability to a well-supported hypothesis while registering high surprise (score) for a less plausible one, reframing hypothesis evaluation as an absolute assessment of intrinsic model predictability.

\subsection{Problem Setup: Confidence as a Plausibility Signal}
\label{sec:overview}
 
For a given paper $p$, let $b_p$ denote its background survey, $q_p$
its research question, and $\mathcal{H}_p = \{h_p^{(1)}, \dots,
h_p^{(16)}\}$ a set of sixteen candidate hypotheses, exactly one of
which, $h_p^{\star}$, is the gold hypothesis, with the remaining
fifteen serving as plausible but incorrect distractors. Rather than
presenting $\mathcal{H}_p$ jointly and requesting a single selection,
we evaluate each candidate independently: for every $h_p^{(i)} \in
\mathcal{H}_p$, an open-weight language model is conditioned on
$(b_p, q_p)$, and its next-token predictions are compared against the
tokens of $h_p^{(i)}$. Because all candidates for a given paper are
scored under an identical background and research question, differences
in score are attributable to the hypothesis text alone, and the
candidate assigned the lowest predictive surprisal is treated as the
model's top choice. Hypothesis evaluation is thereby reduced to a
ranking problem requiring a single forward pass per candidate, with no
explicit comparative judgment step.

\subsection{Prompt Construction and Hypothesis-Span Isolation}
\label{sec:prompt}
 
Each candidate hypothesis is scored using a fixed prompt template that
places the paper's background and research question first, followed by
the candidate hypothesis:
 
\begin{findingbox2}
\footnotesize
Background:\\
{[background survey]}\\[4pt]
Research Question:\\
{[research question]}\\[4pt]
Hypothesis:\\
{[candidate hypothesis text]}
\end{findingbox2}
 
The full prompt is tokenized and processed in a single forward pass,
producing next-token logits at every position. Only positions
corresponding to the hypothesis span are used in scoring: the
background and research question are identical across all sixteen
candidates for a given paper, so including their token-level
predictions would add shared noise without aiding discrimination among
candidates. The hypothesis span is isolated by separately tokenizing
the fixed prefix (everything up to and including the literal string
``\texttt{Hypothesis:}'') and using its token length as an offset
into the full tokenized sequence. Token positions at or beyond this
offset constitute the hypothesis span $\mathcal{H}$ and contribute to
the score; positions preceding it are discarded. This ensures that the
resulting score reflects the model's assessment of the hypothesis text
alone, with the conditioning context held fixed.
 
\subsection{Logit-Based Energy Scores}
\label{sec:likelihood-scoring}
 
Let $x = (x_1, \dots, x_n)$ denote the tokenized prompt, with the
hypothesis span occupying the final positions $t \in \mathcal{H} = \{s,
\dots, n\}$, where $s$ is the offset determined in
Section~\ref{sec:prompt}. Let $z_t \in \mathbb{R}^{|V|}$ denote the
model's output logit vector at position $t-1$ (i.e., the logits used to
predict token $x_t$), where $V$ is the model's vocabulary. From $z_t$
we derive two complementary token-level scores, one normalized and one
unnormalized, and aggregate each over the hypothesis span into a single
scalar per candidate.
 
\subsubsection{Softmax Negative Log-Likelihood (NLL) Energy Score}
The first score is the standard per-token cross-entropy between the
model's predictive distribution and the token observed at that
position:
\begin{equation}
\ell_t \;=\; -\log \operatorname{softmax}(z_t)[x_t]
\;=\; -\log \frac{\exp(z_t[x_t])}{\sum_{v \in V} \exp(z_t[v])}.
\label{eq:nll}
\end{equation}
$\ell_t$ is small when the model's normalized probability distribution
concentrates mass on the observed token $x_t$, and large otherwise.
Because $\ell_t$ is computed after softmax normalization, it reflects
the model's calibrated probability estimate for the hypothesis text; a
lower NLL indicates that the model's predictive distribution assigns
higher likelihood to the hypothesis given the context, which we
interpret as evidence of plausibility.
 
\subsubsection{Raw Target-Logit Energy Score}
The second score follows the general framing of unnormalized
log-probabilities as an energy function over the vocabulary. For each
position in the hypothesis span, the energy is defined as the negated
raw logit assigned to the observed token, prior to softmax
normalization:
\begin{equation}
E_t \;=\; -\, z_t[x_t].
\label{eq:energy}
\end{equation}
Unlike $\ell_t$, $E_t$ is not renormalized against the remainder of the
vocabulary distribution and is therefore directly sensitive to the
scale and calibration of the model's output logits: two hypotheses with
identical softmax probability can receive different energies if the
underlying logit distribution differs in sharpness. This score is
computed in addition to NLL to test whether ranking conclusions are an
artifact of softmax normalization or hold under a complementary,
unnormalized notion of confidence.
For brevity, we refer to the softmax negative log-likelihood score as \textit{NLL} and the raw target-logit energy score as \textit{Raw} throughout the remainder of the paper, including in all tables and figures. Finally, note that the two criteria differ only in the partition term, which measures the model's total confidence mass at that position independent of the observed token. NLL divides it out; Raw retains it. 
\subsubsection{Sequence-Level Aggregation}
Both token-level scores are aggregated over the hypothesis span by
averaging, yielding one scalar per candidate hypothesis $h$:
\begin{equation}
S_{\mathrm{NLL}}(h) \;=\; \frac{1}{|\mathcal{H}|} \sum_{t \in \mathcal{H}} \ell_t,
\qquad
S_{Raw}(h) \;=\; \frac{1}{|\mathcal{H}|} \sum_{t \in \mathcal{H}} E_t.
\label{eq:aggregate}
\end{equation}
(We additionally report the unnormalized \emph{sum} over $\mathcal{H}$
for both scores, to check sensitivity to hypothesis length; the mean in
Eq.~\eqref{eq:aggregate} is our primary score.) Both $S_{\mathrm{NLL}}$
and $S_{Raw}$ are \textbf{lower-is-better}: for a given paper $p$, we rank
all sixteen candidates in $\mathcal{H}_p$ in ascending order of score
and record the rank $r_p$ of the gold hypothesis $h_p^{\star}$ as the
outcome of interest for that paper. A model that reliably assigns the
gold hypothesis the lowest energy achieves $r_p = 1$; our top-line
metric across the benchmark is the fraction of papers for which
$r_p = 1$.

\subsection{Scoring Pipeline}
\label{sec:algorithm}
 
Algorithm~\ref{alg:scoring} summarizes the procedure applied to every
(paper, candidate) pair. The prefix is tokenized separately from the
full sequence (lines~\ref{line:offset}--\ref{line:span}) to locate the
hypothesis span, ensuring that the shared background and research
question never contribute to the score for any candidate.

\begin{algorithm}[t]
\caption{Logit-Based Energy Scoring for One Paper}
\label{alg:scoring}
\begin{algorithmic}[1]
\Require background $b_p$, research question $q_p$, candidate set
$\mathcal{H}_p = \{h_p^{(1)}, \dots, h_p^{(16)}\}$, model $M$
\Ensure rank $r_p$ of the gold hypothesis under $S_{\mathrm{NLL}}$ and $S_{Raw}$
\For{each candidate $h \in \mathcal{H}_p$}
    \State prefix $\gets$ \Call{FormatPrompt}{$b_p$, $q_p$} \Comment{everything through ``Hypothesis:''}
    \State $s \gets |\Call{Tokenize}{\text{prefix}}|$ \label{line:offset} \Comment{hypothesis-span offset}
    \State $x \gets \Call{Tokenize}{\text{prefix} \,\|\, h}$
    \State $\mathcal{H} \gets \{s, \dots, |x|\}$ \label{line:span} \Comment{token positions belonging to $h$}
    \State $\{z_t\}_{t=1}^{|x|} \gets \Call{Forward}{M, x}$ \Comment{single forward pass, teacher forcing}
    \State $S_{\mathrm{NLL}}(h) \gets \frac{1}{|\mathcal{H}|}\sum_{t \in \mathcal{H}} -\log \operatorname{softmax}(z_t)[x_t]$
    \State $S_{Raw}(h) \gets \frac{1}{|\mathcal{H}|}\sum_{t \in \mathcal{H}} -z_t[x_t]$
\EndFor
\State Rank $\mathcal{H}_p$ ascending by $S_{\mathrm{NLL}}$ (resp.\ $S_{Raw}$)
\State $r_p \gets$ rank position of the gold hypothesis $h_p^{\star}$
\State \Return $r_p$
\end{algorithmic}
\end{algorithm}

Scoring a candidate requires only a single forward pass, with no
sampling and no additional model calls; the procedure is therefore
deterministic given the model weights, and the cost of evaluating a
paper scales linearly in the number of candidates (i.e., $16$).

\subsection{Proprietary LLM-as-Judge}
\label{sec:baseline}
As a point of comparison, we evaluate a proprietary, instruction-tuned model accessible only through an API, for which token-level logits are not exposed and the logit-based scoring pipeline described above cannot be applied. For this baseline,  we construct a single zero-shot prompt for each paper as shown
below. We restrict the prompted model to identifying its top 5 candidates rather than producing a full ordering of all 16,
reasoning that selecting the most plausible subset is a simpler and more tractable task for the model than exhaustively
ranking the entire pool.
This baseline represents standard practice in using a large commercial model as an explicit judge and enables a direct comparison between comparative
judgment and intrinsic per-candidate confidence as signals for hypothesis evaluation. The following prompt template is used for proprietary LLMs.

\begin{findingbox2}
\footnotesize
You are a scientific research assistant. Your task is to evaluate a set of hypotheses for a given research question and background, then identify the top 5 most plausible and well-supported hypotheses.\\[4pt]
\#\# Background\\
{[background survey]}\\[4pt]
\#\# Research Question\\
{[research question]}\\[4pt]
\#\# Hypotheses\\
{[indexed list of all 16 candidate hypotheses]}\\[4pt]
\#\# Instructions\\
- Carefully read each hypothesis.\\
- Rank all 16 hypotheses from most to least plausible given the background and research question.\\
- Return ONLY a JSON object in this exact format, with no extra text:\\[4pt]
\{"ranking": [rank-1 index, rank-2 index, rank-3 index, rank-4 index, rank-5 index]\}\\[4pt]
The indices must be integers from 0 to 15. Return only the top 5.
\end{findingbox2}
 
 
\begin{figure}
    \centering
    \includegraphics[width=1\linewidth]{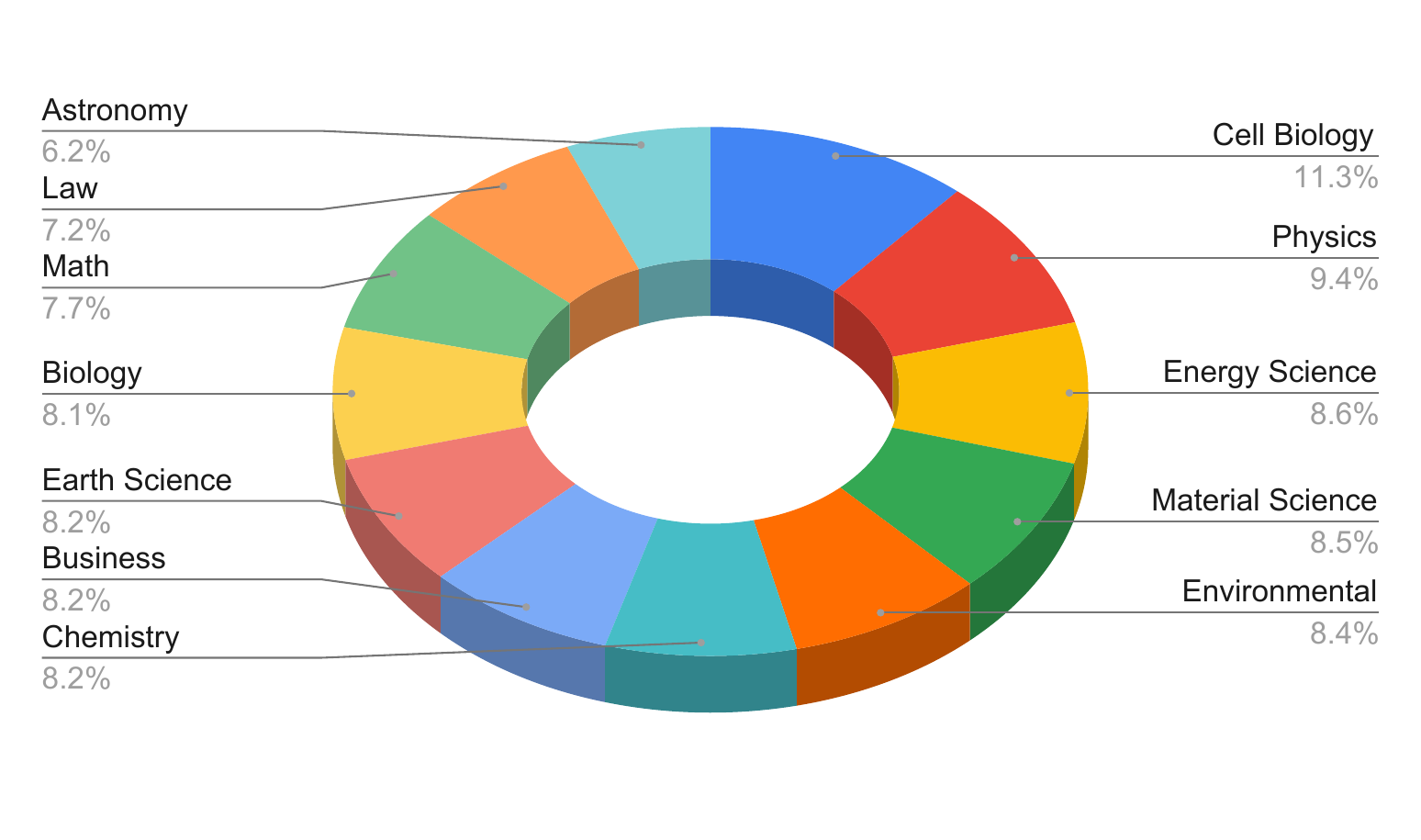}
    \caption{Distribution of papers across disciplines in ResearchBench \cite{liu2026researchbench}.}
    \label{fig:researchbench_disciplines}
\end{figure}

\section{Experimental setup}
\subsection{Dataset}
We used ResearchBench\cite{liu2026researchbench} which is constructed from 1,323 papers across 12 disciplines (Figure \ref{fig:researchbench_disciplines}). 
Each instance in this dataset corresponds to a single published paper and consists of: (i) a \emph{background survey}, a passage summarizing the prior literature and context motivating the study; (ii) a \emph{research question} posed by the paper; and (iii) a pool of $16$ candidate hypotheses of which exactly one is the \emph{gold hypothesis}. We build directly on this decomposition and formulate hypothesis identification as a within-paper ranking task. Given the background survey and research question, each LLM in our paper produces a complete ranking of the candidate hypotheses. A system's performance is determined by the extent to which it ranks the gold hypothesis above the candidate hypotheses.

\begin{table}[htbp]
\centering
\caption{Models evaluated with knowledge cutoff date, parameter counts, release dates, and license terms.}
\label{tab:models}
{\fontsize{6}{6}\selectfont
\renewcommand{\arraystretch}{1.3}
\begin{tabular}{@{}lllr@{}}
\toprule
\textbf{Model} & \textbf{Cutoff Date} & \textbf{Params} & \textbf{Release Date (License)} \\
\midrule
Llama 3.2 1B \cite{meta2024llama32_1b_instruct}
& Dec 2023
& 1B
& Sep 2024 (\href{https://www.llama.com/llama3_2/license/}{CL}) \\

Llama 3.2 3B \cite{meta2024llama32_3b}
& Dec 2023
& 3B
& Sep 2024 (\href{https://www.llama.com/llama3_2/license/}{CL}) \\

Gemma 2 2B \cite{gemmateam2024gemma2improvingopen}
& April 2024
& 2B
& Jun 2024 (\href{https://ai.google.dev/gemma/terms}{ToU}) \\

Gemma 4 12B \cite{gemmateam2026gemma4technicalreport}
& Jan 2025
& 12B
& Jun 2026 (\href{https://www.apache.org/licenses/LICENSE-2.0}{Apache 2.0}) \\

Mistral 7B \cite{jiang2023mistral7b}
& Sep 2023
& 7B
& Sep 2023 (\href{https://www.apache.org/licenses/LICENSE-2.0}{Apache 2.0}) \\

Phi-4 \cite{abdin2024phi4technicalreport}
& June 2024
& 14B
& Dec 2024 (\href{https://huggingface.co/microsoft/phi-4/blob/main/LICENSE}{MIT}) \\

GPT OSS 20B \cite{openai2025gptoss120bgptoss20bmodel}
& June 2024
& 20B
& Aug 2025 (\href{https://www.apache.org/licenses/LICENSE-2.0}{Apache 2.0}) \\

GPT-5 \cite{singh2026openaigpt5card}
& Azure API
& N/A
& Aug 2025 (\href{https://openai.com/en-GB/policies/service-terms/}{Service Terms}) \\
\bottomrule
\end{tabular}
}
\end{table}
\subsection{Compute Setup}
We report results for $8$ language models, ranging from roughly one to twenty billion parameters (Table \ref{tab:models}). Each model is scored under both the NLL and Raw energy criteria described in Section~\ref{sec:likelihood-scoring}. GPT-5 (proprietary) was evaluated under the zero-shot listwise prompting protocol. All results below are computed over the same pool of $1{,}323$ papers. Open-weight models were run on a shared compute node with NVIDIA A100-SXM4 GPUs (80 GB memory each), two AMD EPYC 7742 64-core processors, and 2 TB of system memory, running Ubuntu 22.04.5. We used Python 3.12.13, PyTorch 2.6.0 with CUDA 12.4, and Transformers 5.12.1. The proprietary model was accessed through the Azure API and used no local GPUs.
\subsection{Evaluation Metrics}
\label{sec:metrics}
For every paper and every LLM, we obtain the rank $r_p$ of the gold hypothesis within the full sixteen-candidate pool. All metrics below are computed over the resulting distribution of gold-hypothesis ranks across the $P = 1{,}323$ papers in the benchmark.
\begin{itemize}
\item \textbf{Hit@k.} The fraction of papers for which the gold hypothesis is ranked at or above position $k$, for $k \in \{1, 2, 3, 5\}$:
\begin{equation}
\mathrm{Hit}@k = \frac{1}{P} \sum_{p=1}^{P} \mathbb{1}[\, r_p \le k \,]
\label{eq:hitk}
\end{equation}
Hit@1 corresponds to the gold hypothesis being identified as the single most plausible candidate.

\item \textbf{Mean Reciprocal Rank (MRR).} The mean, across papers, of the reciprocal of the gold hypothesis's rank:
\begin{equation}
\mathrm{MRR} = \frac{1}{P} \sum_{p=1}^{P} \frac{1}{r_p}
\label{eq:mrr}
\end{equation}

\end{itemize}
\section{Results}

\subsection{Overall Hypothesis-Identification Performance}
\begin{figure}
    \centering
    \includegraphics[width=1\linewidth]{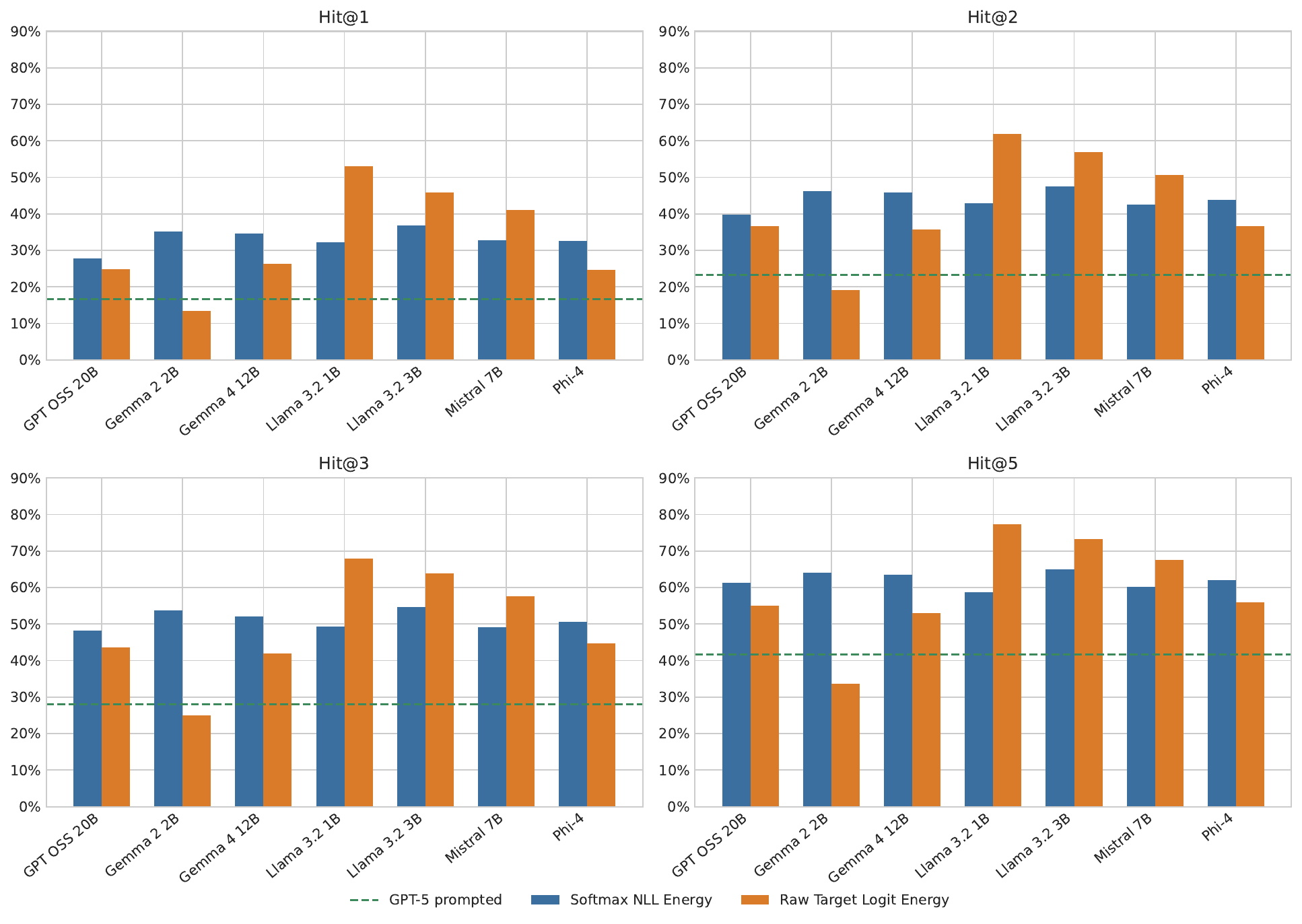}
\caption{\textbf{Hit@k by scoring paradigm and open-weight model.} Each panel shows Hit@$k$ (gold hypothesis ranked in the top $k$ of 16 candidates) for $k \in \{1,2,3,5\}$, across $N=1{,}323$ papers. Bars compare NLL and energy-based scoring per model; the dashed line marks the proprietary model under zero-shot prompted ranking. All panels share the same y-axis scale.}
\label{fig:hitk_grid}
\end{figure}
Table~\ref{tab:overall} reports Hit@$k$ and MRR for every model under every scoring paradigm, aggregated across all papers and disciplines. Reported values are means of per-hypothesis, with 95\% percentile confidence intervals from 2,000 bootstrap resamples. 

\begin{table*}[htbp]
\centering
\small
\caption{Overall hypothesis-ranking performance, aggregated across all 1,323 papers. GPT-5 reports top-5 ranks and MRR is omitted (see Section V-C).}
\label{tab:overall}
\begin{tabular}{llccccc}
\toprule
Model  & Hit@1 & Hit@2 & Hit@3 & Hit@5 & MRR \\
\midrule

\multicolumn{6}{l}{\textit{Softmax NLL Energy Score}} \\
\midrule

GPT OSS 20B
& \low{0.278}\,{\scriptsize[0.254, 0.303]}
& \low{0.398}\,{\scriptsize[0.372, 0.423]}
& \low{0.482}\,{\scriptsize[0.457, 0.509]}
& \midval{0.614}\,{\scriptsize[0.589, 0.640]}
& \low{0.437}\,{\scriptsize[0.418, 0.458]}
\\

Gemma 2 2B
& \midval{0.352}\,{\scriptsize[0.328, 0.378]}
& \midval{0.463}\,{\scriptsize[0.437, 0.489]}
& \midval{0.537}\,{\scriptsize[0.510, 0.565]}
& \good{0.640}\,{\scriptsize[0.615, 0.666]}
& \midval{0.493}\,{\scriptsize[0.474, 0.515]}
\\

Gemma 4 12B
& \midval{0.346}\,{\scriptsize[0.321, 0.371]}
& \midval{0.459}\,{\scriptsize[0.432, 0.485]}
& \midval{0.521}\,{\scriptsize[0.494, 0.548]}
& \good{0.636}\,{\scriptsize[0.610, 0.661]}
& \midval{0.487}\,{\scriptsize[0.466, 0.508]}
\\

Llama 3.2 1B
& \low{0.322}\,{\scriptsize[0.299, 0.347]}
& \low{0.429}\,{\scriptsize[0.402, 0.456]}
& \low{0.494}\,{\scriptsize[0.467, 0.521]}
& \low{0.587}\,{\scriptsize[0.561, 0.613]}
& \low{0.460}\,{\scriptsize[0.439, 0.481]}
\\

Llama 3.2 3B
& \best{0.368}\,{\scriptsize[0.342, 0.395]}
& \best{0.475}\,{\scriptsize[0.448, 0.501]}
& \best{0.546}\,{\scriptsize[0.520, 0.574]}
& \best{0.649}\,{\scriptsize[0.624, 0.675]}
& \best{0.506}\,{\scriptsize[0.486, 0.528]}
\\

Mistral 7B
& \low{0.327}\,{\scriptsize[0.302, 0.353]}
& \low{0.426}\,{\scriptsize[0.401, 0.454]}
& \low{0.492}\,{\scriptsize[0.466, 0.520]}
& \low{0.602}\,{\scriptsize[0.577, 0.628]}
& \low{0.464}\,{\scriptsize[0.444, 0.486]}
\\

Phi-4
& \low{0.327}\,{\scriptsize[0.302, 0.352]}
& \low{0.439}\,{\scriptsize[0.414, 0.466]}
& \midval{0.506}\,{\scriptsize[0.479, 0.533]}
& \midval{0.621}\,{\scriptsize[0.596, 0.646]}
& \low{0.471}\,{\scriptsize[0.451, 0.492]}
\\

\textit{All models}
& \midval{0.331}\,{\scriptsize[0.312, 0.352]}
& \low{0.441}\,{\scriptsize[0.420, 0.463]}
& \midval{0.511}\,{\scriptsize[0.489, 0.533]}
& \midval{0.621}\,{\scriptsize[0.601, 0.642]}
& \midval{0.474}\,{\scriptsize[0.457, 0.492]}
\\

\midrule
\multicolumn{6}{l}{\textit{Raw Target Logit Energy Score}} \\
\midrule

GPT OSS 20B
& \midval{0.249}\,{\scriptsize[0.227, 0.272]}
& \low{0.367}\,{\scriptsize[0.342, 0.394]}
& \low{0.435}\,{\scriptsize[0.409, 0.462]}
& \low{0.550}\,{\scriptsize[0.522, 0.577]}
& \midval{0.403}\,{\scriptsize[0.384, 0.423]}
\\

Gemma 2 2B
& \low{0.135}\,{\scriptsize[0.116, 0.153]}
& \low{0.192}\,{\scriptsize[0.172, 0.212]}
& \low{0.250}\,{\scriptsize[0.228, 0.272]}
& \low{0.336}\,{\scriptsize[0.312, 0.361]}
& \low{0.260}\,{\scriptsize[0.244, 0.276]}
\\

Gemma 4 12B
& \midval{0.263}\,{\scriptsize[0.239, 0.286]}
& \low{0.357}\,{\scriptsize[0.331, 0.380]}
& \low{0.420}\,{\scriptsize[0.393, 0.446]}
& \low{0.531}\,{\scriptsize[0.504, 0.557]}
& \midval{0.402}\,{\scriptsize[0.382, 0.422]}
\\

Llama 3.2 1B
& \best{0.531}\,{\scriptsize[0.505, 0.558]}
& \best{0.620}\,{\scriptsize[0.594, 0.646]}
& \best{0.679}\,{\scriptsize[0.654, 0.704]}
& \best{0.774}\,{\scriptsize[0.751, 0.797]}
& \best{0.641}\,{\scriptsize[0.620, 0.662]}
\\

Llama 3.2 3B
& \good{0.458}\,{\scriptsize[0.432, 0.485]}
& \good{0.568}\,{\scriptsize[0.544, 0.594]}
& \good{0.639}\,{\scriptsize[0.614, 0.664]}
& \good{0.732}\,{\scriptsize[0.709, 0.757]}
& \good{0.586}\,{\scriptsize[0.566, 0.608]}
\\

Mistral 7B
& \good{0.411}\,{\scriptsize[0.386, 0.439]}
& \good{0.506}\,{\scriptsize[0.480, 0.534]}
& \good{0.576}\,{\scriptsize[0.549, 0.603]}
& \good{0.676}\,{\scriptsize[0.651, 0.702]}
& \good{0.538}\,{\scriptsize[0.517, 0.560]}
\\

Phi-4
& \midval{0.246}\,{\scriptsize[0.224, 0.271]}
& \low{0.367}\,{\scriptsize[0.341, 0.393]}
& \low{0.447}\,{\scriptsize[0.422, 0.474]}
& \low{0.559}\,{\scriptsize[0.534, 0.586]}
& \midval{0.405}\,{\scriptsize[0.386, 0.425]}
\\

\textit{All models}
& \midval{0.328}\,{\scriptsize[0.314, 0.342]}
& \midval{0.425}\,{\scriptsize[0.411, 0.441]}
& \midval{0.492}\,{\scriptsize[0.477, 0.507]}
& \midval{0.594}\,{\scriptsize[0.580, 0.608]}
& \midval{0.462}\,{\scriptsize[0.450, 0.475]}
\\

\midrule
\multicolumn{6}{l}{\textit{Prompted}} \\
\midrule

GPT-5
& \low{0.166}\,{\scriptsize[0.146, 0.186]}
& \low{0.232}\,{\scriptsize[0.209, 0.255]}
& \low{0.280}\,{\scriptsize[0.254, 0.304]}
& \low{0.417}\,{\scriptsize[0.389, 0.443]}
& --

\\
\bottomrule
\end{tabular}
\end{table*}

Two patterns stand out. First, the raw target-logit energy criterion is not uniformly better or worse than NLL: it substantially improves ranking performance for the three small open-weight models (Llama 3.2 1B, Llama 3.2 3B, Mistral 7B) relative to NLL, but degrades performance for Gemma 2 2B and leaves GPT OSS 20B, Gemma 4 12B, and Phi-4 roughly comparable to their NLL-based scores. Gemma 2 applies tanh-based soft-capping to its final logits \cite{gemmateam2024gemma2improvingopen}, compressing precisely the raw logit magnitude that $S_{Raw}$ reads while leaving the normalized quantity $S_{NLL}$ reads intact. Gemma 4 12B, which does not use the same capping, recovers to 0.263 under the Raw criterion. We offer this explanation as a mechanistic conjecture.

The two criteria are indistinguishable (e.g., Hit@1 0.3315 vs. 0.3276; MRR 0.4742 vs. 0.4622), a gap well inside the ±0.026 margin of a single Hit@1 estimate at N = 1,323. The variance across models is far larger than the variance across scorers. The highest single cell is Llama 3.2 1B under the Raw criterion (Hit@1 = 0.5314 [0.504, 0.558]), but as the maximum of 14 model-by-scorer combinations chosen after seeing the results, this value should be read as an upper bound on what a well-matched pairing can achieve rather than as an unbiased estimate of that pairing's performance.
 
Second, GPT-5 under zero-shot listwise prompting performs markedly worse than nearly every open-weight likelihood-based configuration: its Hit@1 of 0.1663 and Hit@5 of 0.4172 fall below every NLL-scored model and every Raw-scored model except Gemma 2 2B under the energy criterion. In other words, asking a large proprietary model to explicitly rank all 16 candidate hypotheses in a single pass is, in aggregate, a less reliable indicator of the true hypothesis than measuring how probable a much smaller open-weight model already finds the hypothesis text to be, without any explicit ranking instruction.

We highlight that ResearchBench \cite{liu2026researchbench} used published papers (in 2024 or later). Hence, memorization is a natural concern. However, in our experiments, the strongest results come from Llama 3.2 1B and 3B, whose training data ends in December 2023, well before these papers were published. While the models with later cutoffs perform worse. This does not rule out memorization, but the pattern runs opposite to what it would predict.

\subsection{Per-Discipline Performance}
 
Table~\ref{tab:discipline} breaks down Hit@1 and Hit@5 by discipline for each of the three scoring paradigms (NLL and Raw pooled across all seven open-weight models, and GPT-5 alone). The best-performing paradigm for each discipline and metric is shown in bold.
 

\begin{table}[t]
\centering
\scriptsize
\setlength{\tabcolsep}{2.5pt}
\caption{Hit@1 and Hit@5 by discipline.}
\label{tab:discipline}
\begin{tabularx}{\columnwidth}{@{}lYYYYYY@{}}
\toprule
& \multicolumn{3}{c}{Hit@1} & \multicolumn{3}{c}{Hit@5} \\
\cmidrule(lr){2-4} \cmidrule(lr){5-7}
Discipline & NLL & Raw & GPT-5 & NLL & Raw & GPT-5 \\
\midrule

Astronomy
& 0.308 & \rawbest{0.326} & 0.195
& \nllbest{0.678} & 0.610 & 0.488 \\

Biology
& 0.379 & \rawbest{0.382} & 0.243
& 0.626 & \rawbest{0.628} & 0.449 \\

Business
& \nllbest{0.439} & 0.427 & 0.130
& \nllbest{0.742} & 0.702 & 0.370 \\

Cell Biology
& \nllbest{0.376} & 0.371 & 0.180
& \nllbest{0.666} & 0.650 & 0.400 \\

Chemistry
& \nllbest{0.345} & 0.332 & 0.065
& \nllbest{0.622} & 0.569 & 0.315 \\

Earth Science
& \nllbest{0.421} & 0.413 & 0.046
& \nllbest{0.720} & 0.664 & 0.185 \\

Energy Science
& \nllbest{0.264} & 0.228 & 0.246
& 0.524 & 0.493 & \gptbest{0.553} \\

Environmental Science
& 0.381 & \rawbest{0.394} & 0.081
& \nllbest{0.740} & 0.678 & 0.324 \\

Law
& 0.298 & \rawbest{0.328} & 0.190
& 0.568 & \rawbest{0.603} & 0.368 \\

Material Science
& 0.188 & 0.196 & \gptbest{0.212}
& 0.435 & 0.441 & \gptbest{0.575} \\

Math
& \nllbest{0.371} & 0.368 & 0.176
& \nllbest{0.646} & 0.633 & 0.422 \\

Physics
& 0.210 & 0.184 & \gptbest{0.224}
& 0.507 & 0.475 & \gptbest{0.544} \\

\bottomrule
\end{tabularx}
\end{table}

Likelihood-based scoring (NLL or Raw) yields the best Hit@1 and Hit@5 in the large majority of disciplines. GPT-5 is nominally best in Physics and Material Science, though per-discipline N (approximately 110 to 125) makes these differences statistically indistinguishable (Table~\ref{tab:discipline}). We note this pattern as a hypothesis for future work rather than a finding.

\begin{table}[t]
\centering
\caption{Percentage of head-to-head (pairwise) comparisons in which the correct hypothesis
outranks a candidate.}
\label{tab:pairwise}
\begin{tabular}{ll}
\toprule
Model &  Gold beats candidate \\
\midrule
\multicolumn{2}{l}{\textit{Softmax NLL Energy Score}} \\
\midrule
GPT OSS 20B    & 70.9\% \\
Gemma 2 2B     & 73.2\% \\
Gemma 4 12B    & 72.7\% \\
Llama 3.2 1B   & 68.8\% \\
Llama 3.2 3B   & 74.3\% \\
Mistral 7B     & 69.8\% \\
Phi-4          & 71.7\% \\
\midrule
\multicolumn{1}{l}{\textit{Raw Target Logit Energy Score}} \\
\midrule
GPT OSS 20B    & 66.7\% \\
Gemma 2 2B     & 45.8\% \\
Gemma 4 12B    & 63.7\% \\
Llama 3.2 1B   & \textbf{82.2\%} \\
Llama 3.2 3B   & 79.1\% \\
Mistral 7B     & 74.7\% \\
Phi-4         & 67.3\% \\
\midrule
GPT-5        & 37.4\% \\
\bottomrule
\end{tabular}
\end{table}
\subsection{Pairwise Comparison of Correct \& Competing Hypotheses}
Table~\ref{tab:pairwise} reports how reliably each method separates a paper's actual hypothesis from the alternatives it is presented alongside. For every paper we take the correct hypothesis and compare it against each competing one in turn, and record the share of these comparisons in which the correct hypothesis is placed higher; a method that ordered hypotheses at random would score 50\%. Likelihood-based scoring is stable across models, clustering between 69\% and 74\% regardless of scale, which suggests the signal it captures is a general property of the models rather than an artifact of any one of them. Raw target-logit energy-based scoring is more variable but reaches the strongest results overall, with Llama 3.2 1B at 82.2\% and Llama 3.2 3B at 79.1\% outperforming every likelihood-based configuration; the exception is Gemma 2 2B
at 45.8\%, which is indistinguishable from chance and indicates that the Raw signal is not informative for that model. Listwise prompting of GPT-5 performs worst at 37.4\%, falling below what random ordering would achieve. 
\subsection{Cross-Model Agreement}
\label{sec:rank-caveat}
 
Across the seven open-weight models, all seven agree in ranking the gold hypothesis first for 167 of 1,323 papers (12.6\%) under NLL scoring, but only 3 of 1,323 papers (0.2\%) under Raw scoring. This indicates that while NLL-based rankings are of similar aggregate quality to (slightly ahead in seven of twelve disciplines and slightly behind in the remainder) raw target logit-based rankings, they are considerably more consistent with one another across models, whereas Raw-based rankings vary more sharply from model to model despite comparable average performance.
 
One further point is needed to correctly interpret Hit@$k$ for the prompted model. Because GPT-5 is asked only to produce an explicit top-5 ranking, all candidates excluded from the top 5 are, by construction, placed in a single unranked tied group; under the fixed tie-breaking convention used to place these candidates on a numerical scale, the gold hypothesis, when excluded from the top 5, is always assigned the same fixed rank rather than a range of ranks reflecting a graded preference. Consistent with this, the empirical rank distribution for GPT-5 has non-zero mass only at ranks 1 through 5 (16.6\%, 6.6\%, 4.8\%, 5.7\%, and 8.0\% of papers, respectively) and at the single terminal rank, which accounts for the remaining 58.3\% of papers. For this reason, Table~\ref{tab:overall} reports only Hit@$k$ for $k \le 5$ for GPT-5 and omits MRR, which would conflate ``ranked just outside the top 5'' with ``ranked as the least plausible candidate of all 16.'' Hit@5 for GPT-5 is therefore equivalent to its raw top-5 inclusion rate (41.7\%, 552 of 1,323 papers) and is the most directly interpretable summary statistic for the prompt-based paradigm.
 
\section{Discussion}
Across all three scoring paradigms tested, a consistent picture emerges. Likelihood-based scoring of open-weight models, whether measured through NLL or through the Raw criterion, substantially outperforms zero-shot listwise prompting of a proprietary flagship model at identifying the correct scientific hypothesis from a pool of sixteen candidates. The strongest single configuration in our results is not the largest or most capable model: it is a one-billion-parameter open-weight model scored under the Raw criterion, which outperforms nearly every other model, scoring function, and paradigm we tested, including a substantially larger proprietary model that was explicitly instructed to rank the candidates. At the same time, the advantage of likelihood-based scoring is not uniform across disciplines: prompted ranking is nominally best in Physics and Material Science and competitive in Energy Science. With roughly 110 to 125 papers per discipline, however, these gaps are within sampling noise, and we report the pattern as a hypothesis worth testing at larger per-discipline N rather than as a finding. And while Raw target-logit energy scoring achieves the best raw performance of any configuration, it is also considerably less consistent across models than NLL-based scoring, with far fewer papers on which every model agrees that the gold hypothesis is the top choice.
 
\subsection{Mechanistic Signal May Outperform Explicit Prompting}
 
The most interesting result of this study is the qualitative contrast it points to: a mechanistic signal, obtained without ever asking a model to reason about or articulate a preference, can rival or exceed the performance of directly instructing a much larger model to do exactly that. NLL and Raw energy scores are computed purely from the model's own next-token distribution over the hypothesis text, conditioned on the background and research question; the model is never told that a ranking task is taking place, never asked to compare candidates against one another, and never required to produce a well-formed structured response. Prompted ranking, by contrast, asks the model to explicitly read all sixteen candidates, reason about their relative merit, and report a preference in a specific format. This is a strictly harder and more indirect pathway to the same judgment: it depends on the model's ability to hold sixteen candidates in context simultaneously, to follow the ranking instruction faithfully, and to externalize a judgment that may or may not track its own internal estimate of plausibility.
 

Our results align with prior work showing that a model’s stated judgments do not always reflect its internal probabilities. Similar gaps have been found for bias: a model’s outputs and internal reasoning can show different bias signals \cite{rajwal2025biasedmodelsbiasedthoughts}. If the same is true here, then for tasks like ours, choosing between a specific hypothesis and plausible alternatives, it may be better to use the model’s likelihoods directly rather than asking it to explain its own uncertainty. We believe this is the key novel aspect of our results: unlike prior work on this benchmark, which compares prompted models, we compare models against their own internal likelihoods.
 
The fact that the smallest model in our study performs best overall is worth noting, although we are cautious about drawing strong conclusions from it. This suggests that a model's usefulness for this task does not simply increase with size. Other factors may matter, such as how the model was trained, what data it saw, or how confidently it assigns probabilities to different answers. We did not measure these factors directly, so we cannot say which one explains the result. We see this as an open question for future work.
 
\subsection{Limitations}
This work is an initial step rather than a definitive evaluation of energy-based scoring, prompted ranking, or their relative strengths. Several limitations should be considered when interpreting the results. Our benchmark is built from published hypotheses and the authors curated the candidate hypothesis after the given paper was published. As a result, the task is closer to identifying a known hypothesis than evaluating truly open-ended scientific reasoning. In addition, gold hypotheses were written by domain experts and may simply be more fluent than the distractors. This raises the possibility that likelihood-based methods capture writing quality as well as scientific plausibility, which our current design cannot fully separate. Relatedly, the low cross-model agreement under energy scoring leaves open whether the models share a common signal tracking scientific correctness or whether each one is picking up different surface cues that happen to favor the gold hypothesis. Our evaluation is also limited in scope. We tested prompted ranking with only one proprietary model and one zero-shot prompting strategy, so different prompting methods or models could produce different results. Likewise, the open-weight models differ in size, training data, tokenizer, and alignment, making them useful for comparison but not for isolating the cause of performance differences. These limitations mean our findings should be viewed as evidence from one experimental setting rather than a general conclusion about the strengths of energy-based scoring or prompted ranking.

\subsection{Future Work}

Our next goal is to evaluate hypothesis ranking on genuinely novel scientific hypotheses. We plan to collaborate with experts across scientific fields to build a benchmark of open research questions paired with unpublished candidate hypotheses \cite{ICLR2025_ea94957d, si2026the}. Experts will independently rank these hypotheses based on scientific judgment, enabling comparison with energy-based and prompt-based methods. Their rankings will also allow us to measure inter-annotator agreement and derive a robust reference ranking. This setting requires a different evaluation protocol: because open research questions lack a single agreed-upon answer, evaluation should focus on agreement between model and expert rankings (e.g., Kendall's $\tau$ or Spearman's $\rho$) rather than retrieval metrics like accuracy/Hit@$k$.
 
\section{Conclusion}
We show that scoring candidate hypotheses by a language model's own likelihood, particularly an energy-based criterion, matches or exceeds explicitly prompting a much larger proprietary model to rank them in aggregate, with per-discipline differences too small at our sample sizes to support paradigm-specific claims. This suggests mechanistic scoring signals deserve more attention relative to prompting-based approaches. Because our benchmark is built from already-published hypotheses with known ground truth, we cannot rule out that likelihood-based scores partly reflect textual familiarity rather than scientific reasoning. We therefore propose evaluating these methods against domain-expert rankings of genuinely novel hypotheses as the critical next step.

\section*{Acknowledgments}
This work was primarily supported by the U.S. Department of Energy, Office of Science, Office of Advanced Scientific Computing Research and Office of Basic Energy Sciences, Scientific Discovery through Advanced Computing (SciDAC) program under the FORUM-AI project. Swati Rajwal gratefully acknowledges support through the Graduate Research at Oak Ridge National Laboratory (GRO) program. Sanjay Das is sponsored by the Office of the Laboratory Director, Oak Ridge National Laboratory’s Operational Excellence Initiatives, which is supported by the United States Department of Energy (DOE)’s Office of Science under Contract No. DE-AC05-00OR22725. 
\section*{Data and Code Availability}
The data used in this study are publicly available from a previously published study \cite{liu2026researchbench}. The code and associated artifacts will be released upon acceptance.

\bibliographystyle{IEEEtran}
\bibliography{references}
\balance

\end{document}